# Establishing Vocabulary Tests as a Benchmark for Evaluating Large Language Models

Gonzalo Martínez [1], Javier Conde [2], Elena Merino-Gómez [3], Beatriz Bermúdez-Margaretto [4], José Alberto Hernández [1], Pedro Reviriego [2], Marc Brysbaert [5]

[1] Departamento de Ingeniería Telemática, Universidad Carlos III de Madrid

[2] Departamento de Ingeniería de Sistemas Telemáticos, Universidad Politécnica de Madrid

[3] Escuela de Ingenierías Industriales, Universidad de Valladolid

[4] Departamento de Psicología Básica, Psicobiología y Metodología de las CC. del Compto, Universidad de Salamanca

[5] Department of Experimental Psychology, Ghent University

Address:   Marc Brysbaert

Department of Experimental Psychology

Ghent University

B-9000 Gent Belgium

Marc.brysbaert@ugent.be

**Abstract**

Vocabulary tests, once a cornerstone of language modeling evaluation, have been largely overlooked in the current landscape of Large Language Models (LLMs) like Llama, Mistral, and GPT. While most LLM evaluation benchmarks focus on specific tasks or domain-specific knowledge, they often neglect the fundamental linguistic aspects of language understanding and production. In this paper, we advocate for the revival of vocabulary tests as a valuable tool for assessing LLM performance. We evaluate seven LLMs using two vocabulary test formats across two languages and uncover surprising gaps in their lexical knowledge. These findings shed light on the intricacies of LLM word representations, their learning mechanisms, and performance variations across models and languages. Moreover, the ability to automatically generate and perform vocabulary tests offers new opportunities to expand the approach and provide a more complete picture of LLMs' language skills.

**Latent semantic analysis as a way to understand word meaning and word acquisition**

Machine language learning goes back to a seminal paper by Landauer and Dumais (1997), who presented latent semantic analysis (LSA) as a new theory of knowledge representation. Meaning was inferred from local co-occurrences of words in representative text. The main idea was that one can learn the meaning of an unfamiliar word X from the words that frequently occur with X. Results with the LSA model suggested that English vocabulary could be acquired in that way at a rate comparable to schoolchildren, without prior linguistic or perceptual knowledge.

Landauer and Dumais (1997) not only proposed the theory but also presented a mathematical model that, starting from the co-occurrences of words in the texts, constructed a matrix that was then mapped to a space of reduced dimensions. In this space of a few hundred dimensions, each word was represented by a point and the distance between the points represented the distance in meaning. The points were defined as semantic vectors and had about 300 dimensions. Words with similar meanings had semantic vectors that were close to each other.

Landauer and Dumais (1997) evaluated their LSA model with a vocabulary test. They used 80 items from the synonymy section of the Test of English as a Foreign Language (TOEFL). In this task, target words were presented with four response alternatives from which the correct one had to be chosen. Landauer and Dumais' (1997) found that the semantic vector of the correct answer was closest to that of the target word in 51 items (64% correct). This was comparable to US college candidates from non-English-speaking countries who scored 65% correct on average.

Although the scores of the LSA model were not perfect (they were at the same level as speakers of English as a second language), they were much better than what was available at the time. Jarmasz and Szpakowicz (2003) reported that other algorithms for word meaning scored barely above chance level (25%) on the TOEFL test.

In subsequent years, authors tried to improve the performance of LSA-type models by optimizing the algorithm and training materials. Performance on the TOEFL test (as it became known) gradually increased until Bullinaria and Levy (2012) reported 100% correct performance for a model tweaked to the test (ACL Wiki, 2019).

Landauer and Dumais (1997) argued that their LSA model could be viewed as a simple three-layer neural network, an argument that was confirmed when real connectionist networks were developed (Mikolov et al., 2010, 2013) that quickly outperformed the co-occurrence statistics used in LSA (Mandera et al., 2017).

The work of Landauer and Dumais (1997) showed how vocabulary can be learned from a large corpus of text using only local relationships between words. This breakthrough was a fundamental step in understanding how language can be acquired and how computer systems can be implemented that can learn the meaning of words and text. These concepts were the forerunners of today's Large Language Models (LLMs).

**Large language models design and evaluation**

Over the past decade, artificial intelligence has made impressive progress in language modeling (and other areas). The improvements were possible due to the availability of huge text datasets, more powerful computing and storage, and architectures that can implement very large models efficiently (such as transformers). First the introduction of the transformer (Vaswani 2017) and then the development of popular models based on it, such as BERT (Devlin 2019) or T5 (Raffel 2020), paved the way for the development of LLMs with many billions of parameters, such as GPT4 (Chang et al, 2023). Transformers are complex neural networks with many interconnected layers and novel mechanisms such as attentional focus that allow them to learn complex relationships, for example, between words in text. By increasing the size of transformers and training datasets, unprecedented performance has been achieved in many language processing tasks, but more importantly, they have been integrated into products such as ChatGPT, Bard or Bing reaching hundreds of millions of users (Ray 2023). The main features of these LLMs are the huge size of their training datasets and number of model parameters, their ability to learn different languages (including programming languages) and perform a wide range of tasks such as generating text in genres, translating, answering questions and summarizing (Sallam 2023). A particularly attractive application of LLMs is the development of intelligent chatbots, such as the well-known ChatGPT used by hundreds of millions of people worldwide.

LLMs operate on units called tokens, which in some cases correspond to words but can also be sequences of N letters in the training dataset. The text is decomposed into tokens as input, and the model generates output tokens that are assembled into words and sentences. LLMs are trained to predict the next token in a sentence (Zhao 2023). This is done by taking existing texts, removing tokens from them and using these tokens as criteria for the outcome to be predicted by the model. This approach makes it possible to use huge datasets containing almost every text of interest on the Internet. Interestingly, predicting the next token assumes that languages and words can be learned from the words that appear in texts together with the target word. This is exactly the assumption Landauer and Dumais (1997) made in their LSA model. Therefore, LLMs can be seen as descendants of LSA. The prediction of words based on surrounding words is not the only link between LLMs and LSA. The use of a point in a multidimensional space to represent word meanings, introduced by LSA, is also used by LLMs to map inputs to so-called "embeddings." These embeddings are vectors of values that correspond to a point in the space where the meaning is located, so that, similar to semantic vectors in LSA, points that are close together correspond to similar meanings. In the case of LLMs, embeddings are usually larger than the 300 dimensions used in LSA, with several thousand dimensions.

To evaluate LLM performance, several benchmarks have been proposed (Chang 2023). In most cases, the test evaluates how well LLMs answer questions on almost any topic (Hendrycks 2020; Srivastava 2022) or are able to perform reasoning based on a given text (Zellers 2019). For example, there are benchmarks with thousands of mathematical problems covering almost every discipline in mathematics (Hendrycks 2021). There are also frameworks that can be easily extended so that new tasks or tests can be added at will.[1] Those expanded benchmarks focus on quantifying how well LLMs perform on different knowledge tasks.

---

[1] https://github.com/openai/evals

**Large language models produce language, which is likely to create a feedback loop**

LLMs do not just answer questions or solve problems. At the same time, they produce language output. They create text when they provide answers or translations. In fact, they are already being used to help write entire novels and textbooks. So people and future LLMs will be increasingly exposed to LLMs' output, creating a feedback loop (Martínez 2023a). As a simple example, take a word that is not produced by LLMs because a simpler synonym exists. This word will appear less and less in the language to which future people and LLMs are exposed until it becomes extinct.

Some research is beginning to appear on these subtler, as well as more fundamental, linguistic aspects of the use of LLMs, but it is still very limited compared to the large number of articles evaluating the performance of LLMs on various knowledge tasks. Only a few studies have focused on the linguistic characteristics of the text generated by LLMs. A comparison of the linguistic characteristics of humans and LLMs is presented in Muñoz-Ortiz (2023), who focuses on the analysis of news generated by an open-source LLM. There are also studies that focus on phonological (Toro, 2023) and lexical (Reviriego, 2023) aspects of LLMs.

In this article, we look at how LLMs perform on vocabulary tests, what this tells us about how LLMs learn language, and whether vocabulary tests can be a useful addition for evaluating LLMs.

**Vocabulary tests and large language models**

As discussed at the beginning of this article, the TOEFL test was an important tool for evaluating Landauer and Dumais' (1997) LSA model and its later extensions. Vocabulary tests are also widely used to assess language proficiency in humans (Webb & Nation, 2017). Surprisingly, current LLMs are no longer tested on vocabulary tests, probably because everyone assumes they will be error-free. Given that LSA-type models were already achieving flawless performance in early 2010 and that current LLMs are vastly superior in design and the amount of training material, it seems a waste of time to test them on something as simple as the TOEFL test, which only taps into knowledge known to undergraduates with English as a second language. Indeed, this is what the present authors expected when they used the TOEFL test as the starting point of a study that would use more taxing vocabulary tests.

Vocabulary tests traditionally consist of multiple-choice questions that require participants to choose the correct answer from a number of alternatives (Vermeiren & Brysbaert, 2023). Indeed, this is the format of the TOEFL test. The difficulty of the test then depends on the difficulty of the target words and the number and difficulty of the answer alternatives. By varying these, it is possible to make more demanding vocabulary tests than the TOEFL to see what level various LLMs achieve.

Meara and Buxton (1987) introduced another format for vocabulary tests that may be of particular interest for LLM testing. They presented participants with a list of letter strings and asked them to indicate which words they knew. To prevent participants from selecting all items without knowing them, legal letter strings were added that do not exist as words in English (such as "plound" or "ternace"; these are so-called non-words or pseudowords). Performance was estimated based on both word and non-word performance, so that a participant who selected all items would receive a score of zero. The Yes/No format gained momentum when Lemhöfer and Broersma (2012) published an English language proficiency test for use in psycholinguistic research, which they called LexTALE. The Yes/No format is interesting because the test taker must refrain from choosing the non-words. LLMs are known to tend to present nonexistent information based on word co-occurrences (so-called hallucinations), as discussed in Zhao et al. (2023). So we thought it would be interesting to see how LLMs would perform on tests with non-words.

Considering the evaluation of LLMs, vocabulary tests have a number of features that are of interest:

- First, vocabulary tests are pure language tests with stimuli that are not embedded in an informative context. Thus, vocabulary tests could potentially be used to evaluate LLMs' knowledge of languages. This is an important point since LLMs should perform equally well for all the languages they claim to support (see Petrov et al., 2023, for evidence that this unfortunately is often not the case yet).
- Second, if a model makes errors, these errors can be analyzed to study how LLMs learn a language and to evaluate whether cognitive theories of language acquisition apply to LLMs. For example, early acquired words are often well remembered by humans even though they almost never occur in everyday language use (Stadthagen-Gonzalez et al., 2004). Vocabulary tests allow fine-grained evaluation of LLM, which facilitates analysis and understanding of the mechanisms and algorithms underlying the models.
- Third, how do LLMs interpret non-words and are they able to distinguish them from valid words? Do they follow the same mechanisms as humans (Gatti et al., 2023)? How does the tokenization that LLMs use affect word learning? To what extent does performance depend on the specific question posed to the LLM?
- A fourth important feature of vocabulary tests is that they can be automated, both for test generation and test execution. This is especially true for the Yes/No format. There are software tools that can be used to generate non-words in different languages (Keuleers 2010) and even workflows to generate entire test suites (van Rijn et al., 2023). This makes it possible to generate large-scale vocabulary tests. Similarly, the execution of those tests on different LLMs can be automated (Martínez 2023b), allowing the evaluation of multiple LLMs in multiple languages, without practical limitations on the number of words tested.

These characteristics make vocabulary tests potentially interesting for evaluating LLMs. Below we present an initial assessment of the use of existing vocabulary tests for evaluating LLMs so that we can decide whether or not it is a fruitful approach.

**Evaluation of LLMs in English and Spanish**

To assess the usefulness of vocabulary tests in LLM evaluation, we conducted several vocabulary tests on different LLMs. To have a representative sample of current LLMs, we selected two company-owned, commercial LLM tools: ChatGPT[2] (based on GPT3.5 and GPT4) and Bard[3] (based on PaLM 2), together with two open source LLMs: Llama (Touvron et al., 2023) and Mistral (Jiang et al., 2023).

ChatGPT, developed by OpenAI, is the most popular LLM-based chatbot today and probably the one that has shown the best performance across a variety of tasks. Two versions of ChatGPT (with different numbers of parameters) were tested. Bard is developed by Google and is intended to compete with ChatGPT, so both are good examples of commercial chatbots. Parameters and source code for these LLMs are not available, which makes them less interesting for research purposes because they can be modified overnight without researchers being able to verify what was done. Still, because of their massive use by the public, it is worthwhile to determine their performance at some point. Llama, developed by Meta, is probably the best known open-source LLM right now. Another open-source LLM with good performance is Mistral, developed by a startup of the same name. We tested three versions of Llama, of different sizes, to see to what extent performance improves as network complexity increases.

---

[2] https://openai.com/blog/chatgpt
[3] https://ai.google/static/documents/google-about-bard.pdf

The seven models considered in our evaluation are summarized in Table 1. They range from relatively small models to the largest models that were publicly accessible in October 2023.

Table 1: LLMs considered in the experiment

| LLM | Company | Type | Parameters |
| --- | --- | --- | --- |
| Mistral-7B | Mistral | Opensource | 7.3 billion |
| Llama-7B | Meta | Opensource | 7 billion |
| Llama-13B | Meta | Opensource | 13 billion |
| Llama-70B | Meta | Opensource | 70 billion |
| PaLM 2 (Bard) | Google | Commercial | > 100 billion (non-official) |
| GPT-3.5-turbo (ChatGPT) | OpenAI | Commercial | 175 billion |
| GPT-4 (ChatGPT) | OpenAI | Commercial | > 1 trillion (non-official) |

The tests were run automatically using the Application Programming Interfaces (APIs) of the LLM-based chatbots to create the questions in each test and then produce an excel file with all the answers, as described in Martínez et al. (2023b). The only exception was Bard, whose API was not accessible in Spain at the time of the evaluation. Automation is interesting for running tests at scale, since we evaluated seven LLMs on tests with dozens of questions each. In addition, the use of the API allowed control over LLM parameters, such as temperature, which adjust the variability of answers. During evaluation, LLMs were not given context information and default parameters were used except temperature, which was set to zero if it was controllable to produce deterministic responses. The prompts used to interrogate the chatbots were simple and similar to those used in the human tests (see below). The performance of LLMs can be improved by providing context or more sophisticated prompts that force the LLMs to solve the questions step by step using a chain of thought (Zhao el al, 2023). However, our goal was to understand how LLMs perform in vocabulary tests when presented with the same questions as humans and not to modify the questions or provide additional information to improve the LLMs' answers.

A number of representative vocabulary tests with multiple-choice and yes/no questions were used in the experiment. The details of the tests are summarized in Table 2.

The first test was the TOEFL introduced by Landauer and Dumais (1997). It contains 80 target words with four alternatives to choose from (these are also singles word). The difficulty of the items varies, but the level is adapted to non-English-speaking students who want to study at English-speaking universities. The test cannot be freely shared due to copyright restrictions, but researchers can request access to the stimulus material, if use is strictly limited to research purposes. We were kindly granted access to the stimuli by the LSA research group at the University of Colorado.

The second vocabulary test was the StuVoc test, published by Vermeiren et al. (2023). This test contains three subtests with 50 validated English items each (thus 150 items in total). Items consist of target words in short neutral sentences along with four response alternatives. Unlike the TOEFL, the alternatives can include short descriptions of words. The first two subtests are difficult enough for English-speaking university students. The third subtest is easier and better suited for second-language speakers with high proficiency (Vermeiren & Brysbaert, 2023). The level of the last subtest is similar to the TOEFL, while the first two tests are more demanding.

The third vocabulary test was the Spanish adaptation of StuVoc. Bermúdez-Margaretto & Brysbaert (2022) translated 146 of the English StuVoc items into Spanish and validated them on a group of adult native Spanish speakers. They selected the 80 best items (good distribution of difficulty levels, good correlation between item performance and overall test performance, and a clear transition from unknown to known based on item response theory analysis). The remaining 66 items were considered less interesting for various reasons.

The fourth and fifth tests were yes/no tests. For English, we used LexTALE, proposed by Lemhöfer and Broersma (2012). The test contains 40 English words and 20 non-words. Since the test is aimed at advanced second language speakers, the level is comparable to the TOEFL. Native speakers typically score more than 90% correct on the test (scores obtained after subtracting the % yes answers to non-words).

Finally, we tested the models on a Spanish Yes/No test published by Izura et al. (2014). This test is more comprehensive and more difficult than the English LexTALE because the authors wanted the test to be usable by both native Spanish speakers and second-language speakers (Ferré & Brysbaert, 2017). The test contains 60 Spanish words and 30 non-words.

We presented all vocabulary tests to all LLMs. For the multiple-choice tests we asked: "Answer the option which is the meaning of the following word "squelch": a. suppress b. revive c. acquire d. dispute. Please, first just answer the letter of the option and below your explanation.". For the Yes/No tests we asked "Please answer "Yes" or "No" to the following question: Is X an existing word in English (Spanish)?"

Table 2: Vocabulary tests considered in the experiment.

| Test | Type | Language | Number of items |
| --- | --- | --- | --- |
| TOEFL | Multiple choice | English | 80 |
| StuVoc-Eng | Multiple choice | English | 150 |
| StuVoc-Esp | Multiple choice | Spanish | 80 |
| LexTALE-Eng | Yes/No | English | 60 |
| LexTALE-Esp | Yes/No | Spanish | 90 |

The results obtained for the multiple-choice tests are summarized in Table 3.

As expected, most models performed well on the TOEFL test, and the larger Llama models outperformed the basic 7B version. Still, performance was not flawless. Moreover, the models differed in the items they got wrong, suggesting that suboptimal performance was not due to one or two weak items. The three commercial models failed on the item "fashion," where they chose the option "rage" instead of "manner" (which Llama-7B and Mistral-7B did get right). For the item "figure," GPT3.5, Bard and Mistral-7B chose the option "express" instead of "solve." GPT4 and Mistral-7B obtained the best scores. However, the differences with PaLM 2 (Bard), Llama-70B and GPT3.5 were small. Unfortunately, no further information can be given for the TOEFL test as the items are copyrighted. For the other tests, the results for each model and item are available in a public GitHub repository.[4]

Despite the fact that the StuVoc-Eng is more demanding for human speakers than the TOEFL, LLMs' performance on this test was generally higher than on the TOEFL. One reason could be the availability of a short, neutral context sentence (given information about the part-of-speech). Another reason could be that multi-word descriptions describe the meaning of the target words better than one-word synonyms. Again, the best scores were obtained by GPT4 and Mistral-7B, closely followed by PaLM 2 and GPT3.5. The Llama models had lower scores and Llama-13B performed better than Llama-70B. The item "Let's not pussyfoot around" was censored by PaLM 2 and Llama-7B. What is further striking is that no item was missed by all LLMs. Every item was scored correctly by at least 4 of the 7 LLMs. Three LLMs had only two items wrong: "The boy shuddered." where they selected "almost fell" instead of "shook" and for "She parried the comments" where the three failing models selected different answers.

There was a significant performance drop for StuVoc-Esp compared to StuVoc-Eng, even though the items were direct translations. The drop was especially large for the open-source models (Llama and Mistral), which fell below 80%. For the commercial models, the decline was smaller, but still noticeable. Bard (PaLM 2) achieved the highest score in this test.

Table 3: Performance of October 2023 LLMs on the multiple-choice tests (percentage of correct answers)

| Test | Llama-7B | Llama-13B | Llama-70B | Mistral-7B | GPT3.5 | GPT4 | PaLM 2 |
|---|---|---|---|---|---|---|---|
| TOEFL | 76.3% | 93.9% | 96.3% | 98.8% | 96.3% | 98.8% | 97.5% |
| StuVoc-Eng | 91.3% | 96.0% | 94.7% | 99.3% | 97.3% | 100% | 98.7% |
| StuVoc-Esp | 73.8% | 76.4% | 61.3% | 68.8% | 93.8% | 95.0% | 96.3% |

The results for the Yes/No tests are presented in Table 4 and are reported independently for words and non-words to better understand the results. Performance on LexTALE-Eng was quite good and tended to be better for words than for non-words. Remember that this is a fairly easy test designed for second language speakers. Interestingly, for Llama we see better performance on words as the model gets larger and at the same time worse performance on non-words. ChatGPT4 (GPT4) was flawless on all items, but Google Bard (PaLM 2) in its tested version was poor (together with Llama-70B) in hallucinating the existence of non-words.

Analyzing the data from LexTALE-Esp, we saw that two of the non-words (vegada, capillo) are existing words in Spanish because they appear in some dictionaries. Therefore, we excluded these two items from the analyses. Performance was good for the words (especially considering that some were more difficult than the LexTALE-Eng words). When further asked about the meaning of the words, most models gave the English translation. However, performance was very poor for the Spanish non-words,

---
[4] https://github.com/WordsGPT/LLM_Vocabulary_Evaluation

where the models not only gave "yes" answers, but when asked, readily gave meanings and translations for letter combinations that do not exist in Spanish. This was especially true for Llama and Mistral. Mistral was unable to identify even one non-word and performed like a boastful test taker, claiming to know all the "words" but in reality scoring zero points. Performance was best for GPT4, but even this model presented interpretations for 18% of the non-words. Performance was also poor for Bard, where interpretations were given for more than half of the non-words.

Table 4: Performance of October 2023 LLMs on the Yes/No tests (percentage of correct answers)

| Test | Type | Llama-7B | Llama-13B | Llama-70B | Mistral-7B | GPT3.5 | GPT4 | PaLM 2 |
|---|---|---|---|---|---|---|---|---|
| LexTALE-En | Words | 92.5% | 87.5% | 100% | 100% | 100% | 100% | 100% |
| LexTALE-En | Non-words | 100% | 95% | 85% | 95% | 95% | 100% | 85% |
| LexTALE-Esp | Words | 96.7% | 96.7% | 100% | 100% | 96.7% | 100% | 100% |
| LexTALE-Esp | Non-words | 7.1% | 60.7% | 14.3% | 0% | 67.9% | 82.1% | 46.4% |

The Spanish results suggest that current LLMs are likely to perform less well in languages other than English, for which the training materials were most extensive. To better understand this effect, the information on the training datasets for GPT3[5], Palm (Chowdhery 2023) and Llama (Touvron et al. 2023) is summarized in Table 5. For the other models the information on the training dataset is not publicly available as it is considered a key element of the design, and it is kept confidential for commercial reasons[6].

What is first noticeable is the small percentage of training materials in Spanish, even though Spanish is one of the most widely spoken languages in the world (and present in the U.S.). To some extent, it is surprising that performance on StuVoc-Esp and for some LLMs on LexTALE-Esp was so good, given the limited amount of Spanish training material these models received. At the same time, the results clearly show that the use of LLMs for languages other than English is oversold.

Table 5: Percentage of the training dataset in English and Spanish for several LLMs

| Model | English | Spanish |
|---|---|---|
| GPT3 | 92.64% | 0.77% |
| PaLM 2 | 77.98% | 2.11% |
| Llama | 89.70% | 0.13% |

**Discussion**

In the early days of machine language models, developers tested the quality of their models with a vocabulary test (specifically, the TOEFL test introduced by Landauer & Dumais, 1997). At present this is no longer done, possibly because developers assume that current models are error-free given the improvements in design and amount of training over the past decade.

---
[5] https://github.com/openai/gpt-3/blob/master/dataset_statistics/languages_by_word_count.csv
[6] https://huggingface.co/mistralai/Mistral-7B-v0.1/discussions/8

When we tested the performance of available LLMs, however, we saw that most models did not achieve 100% scores on the various vocabulary tests (the exception was GPT4 on StuVoc-Eng and LexTALE-Eng). Performance was especially poor in Spanish and in Yes/No tests.

The origin of the poor performance in Spanish is not difficult to find. Given the small percentage of training material the models received in Spanish (Table 5), it is to some extent surprising that the models still performed so well. Given that Spanish is one of the most widely spoken languages in the world (and is present in the U.S.), it is to be expected that performance will decline even further for languages with less training material. It is also to be feared that the texts produced by these models will be of low quality (cf. the issue of the feedback loop). Our findings indicate that training datasets should be more balanced to avoid bias against languages other than English, and they demonstrate the usefulness of vocabulary tests to perform comparisons of language proficiency in LLMs.

The high number of non-words accepted as meaningful by LLMs (again, particularly in Spanish) is also important and worthy of further investigation. Needless to say, language models that hallucinate meaning when there is none are poor assistants. One possible cause could be that some models do not work with words as input and output units, but with tokens of a certain length, regardless of word boundaries. In such models, words have no special status and non-words similar to existing words (as good pseudowords should be) can activate lexical output. Even in humans, there is evidence that some non-words activate word-related meanings (Gatti et al., 2023). Further factors contributing to hallucinations of non-words are likely to be cross-language contamination (a non-word in one language may be a word in another language) and spelling errors in training materials.

Another question is whether the performance of the best English models is already good enough. Several reviewers noted that the performance of these models is already comparable and possibly superior to humans. Much here depends on what one wants (or claims) to achieve. If average human performance is the goal, then the current best models may already be good enough. However, if models are intended to improve human performance, then further progress still seems possible, even in English.

Perhaps even more important than the results with the specific tests we used is the usefulness of vocabulary tests in general to examine the performance of LLMs. Many hypotheses in cognitive science about human cognition can be related to the generation of specific words (and non-words) that can be tested in LLMs, to see if the predicted accuracy differences are obtained. This will likely lead to new theoretical developments that may be applicable to both LLMs and humans. By focusing on words rather than knowledge areas, vocabulary tests provide a fine-grained mechanism for investigating the fundamental cognitive mechanisms of language processing in LLMs (and humans). The vocabulary tests of Table 2 included words that were good for testing language proficiency in general, but words (and non-words) can also be selected to answer specific theoretical questions.

Another aspect that can be varied is the format of the vocabulary test. In this article, we have discussed the multiple-choice format and the yes/no format. Other formats are those in which a correct definition or word must be generated (production rather than recognition). Such tests are more difficult for humans (Zhang & Zhang, 2022), but they may be easier for LLMs. Even within a format, a few changes can make a difference. In the results section, we wondered to what extent the availability of neutral carry phrase improves performance. This can be easily tested in LLMs and, if deemed interesting, later in humans. Similarly, we can examine the extent to which LLMs' performance on Yes/No tests depends on the questions asked and what this says about the underlying mechanisms.

Finally, it is possible to automate the production of large-scale vocabulary tests (e.g., van Rijn et al., 2023). This makes it possible to generate large numbers of stimuli based on different criteria, such as frequency in text, length, or even how letter sequences are tokenized by LLMs. In this way, vocabulary tests can be developed to evaluate LLMs at scale. This is particularly interesting because, unlike human participants, LLMs are not limited in the number of items they can process. Thousands of items can be tested in LLMs, allowing comprehensive evaluation of the entire vocabulary. This also solves the problem of experimenter bias in the selection of stimuli (Forster, 2000; Kuperman, 2015). The evaluation of the results can also be automated as part of the pipeline.

**Conclusion**

We hope to have convinced readers that the development of vocabulary tests for the evaluation of LLMs is an interesting area of research at the intersection of cognitive science, psycholinguistics, and artificial intelligence, which can provide valuable insights into both the operation of LLMs and theories of human cognition. Therefore, there is a strong case for designing such vocabulary tests to complement existing benchmarks for evaluating LLMs.

At the same time, we would like to point out that the performance of LLMs is also useful for creators of vocabulary tests for human participants. The models may not replace item selection based on psychometric analysis, but they can point to ambiguities in item construction. To avoid being overly influenced by one model, our data suggest that it is better to test the performance of multiple models. For the good items we tested, most models gave the expected result, at least as long as the items were existing words. LLM tests can thus be used to avoid problems in new tests for humans. Also note that LLM testing alerted us to the fact that two of the non-words in the Spanish Yes/No test were present in some Spanish dictionaries.

**Availability of the data**

Item-level performance for all items we can share is available on a public repository[7], along with the specific questions asked of LLMs and the code used to run the test on the different LLMs.

**Acknowledgements**

We thank the LSA research group at Colorado University and in particular Peter Foltz for kindly providing the items for the TOEFL test. This work was supported by the FUN4DATE (PID2022-136684O7B-C21/22) project funded by the Spanish Agencia Estatal de Investigación (AEI) 10.13039/501100011033 and by the OpenAI Research Access Program.

---

[7] https://github.com/WordsGPT/LLM_Vocabulary_Evaluation